\documentclass{article}
\pdfoutput=1
\PassOptionsToPackage{numbers, compress}{natbib}

     \usepackage[preprint]{neurips_2020}

\usepackage[utf8]{inputenc} 
\usepackage[T1]{fontenc}    
\usepackage{hyperref}       
\usepackage{url}            
\usepackage{booktabs}       
\usepackage{amsfonts}       
\usepackage{nicefrac}       
\usepackage{microtype}      
\usepackage{amsmath}
\usepackage{wrapfig}
\usepackage{graphicx}
\usepackage{subcaption}
\usepackage{todonotes}

\title{Density-embedding layers: a general framework for adaptive receptive fields}

\author{%
  Francesco Cicala\\
  Department of Mathematics and Geosciences\\
  University of Trieste\\
  francesco.cicala00@gmail.com\\
    \and 
  \textbf{Luca Bortolussi}\\
  Department of Mathematics and Geosciences\\
  University of Trieste\\
  lbortolussi@units.it
}

\begin{document}

\maketitle

\begin{abstract}
    
  The effectiveness and performance of artificial neural networks, particularly for visual tasks, depends in crucial ways on the receptive field of neurons. The receptive field itself depends on the interplay between several architectural aspects, including sparsity, pooling, and activation functions. In recent literature there are several ad hoc proposals trying to make receptive fields more flexible and adaptive to data. For instance, different parameterizations of convolutional and pooling layers have been proposed to increase their adaptivity. In this paper, we propose the novel theoretical framework of density-embedded layers, generalizing the transformation represented by a neuron. Specifically, the affine transformation applied on the input is replaced by a scalar product of the input, suitably represented as a piecewise constant function, with a density function associated with the neuron. This density is shown to describe directly the receptive field of the neuron. Crucially, by suitably representing such a density as a linear combination of a parametric family of functions, we can efficiently train the densities by means of any automatic differentiation system, making it adaptable to the problem at hand, and computationally efficient to evaluate. This framework captures and generalizes recent methods, allowing a fine tuning of the receptive field. In the paper, we define some novel layers and we experimentally validate them on the classic MNIST dataset.
\end{abstract}

\section{Introduction}

Convolutional neural networks (CNN) are a standard architecture for working on tasks involving signals, in particular for visual tasks. They provided numerous state-of-the-art results on popular benchmarks \cite{He2016DeepRL, He2015DelvingDI,Krizhevsky2012ImageNetCW,  Szegedy2015GoingDW, Tan2019EfficientNetRM}, and they continue to receive a lot of interest because of their ability to learn on complex tasks with much less parameters than required by a fully-connected network. The convolutional layer has the property to make an efficient use of its shared weights by means of sparse interactions with the input signal. Every neuron of a convolutional layer will apply an affine transformation only to a local region of the input, and these layers are arranged in a hierarchical structure \cite{LeCun2010ConvolutionalNA}, as supported by the neuroscientific study on visual cortex \cite{Hubel1962ReceptiveFB, Hassabis2017NeuroscienceInspiredAI, Yan2020RevealingFS, Ukita2018CharacterizationON}. Convolutional layers are often combined with pooling layers, which allow to reduce the signal dimensionality while preserving its relevant features.

The effectiveness of this range of methods can be interpreted in light of their receptive field. The receptive field of a neuron with respect to an input signal corresponds to the region of the signal which will affect the output of the neuron \cite{Le2017WhatAT}, i.e. the region whose variation will induce a variation in the neuron's output. Convolutional and pooling layers make use of a convenient reshaping of the neurons' receptive field, and they take advantage of some general properties of temporal and spatial signals (i.e., time series and images). Specifically, the max pooling is supported by biologically-inspired arguments \cite{Riesenhuber1997JustOV, Serre2010ANA}, and it has an established importance in improving the performance of convolutional architectures \cite{Boureau2010ATA}. Moreover, many methods have been proposed to increase the flexibility of these layers \cite{Saeedan2018DetailPreservingPI, Lee2018GeneralizingPF, Lee2019LearningRF, Kobayashi2019GaussianBasedPF, Kobayashi2019GlobalFG, Han2018OptimizingFS}. These methods parameterize the underlying receptive field so that it can adapt to data.

The receptive field has a fundamental relevance in determining the performance of neural networks on visual tasks, since the output must be responsive with respect to a large enough area of the input. It has been noted that size is not a sufficient measure of the receptive field of a neuron, since in general it will not be uniform. In fact, it has been shown in \cite{Luo2016UnderstandingTE} that it has a gaussian distribution, and the effective receptive field is much smaller than the theoretical one.

In this work we observe how, despite the important results that have been obtained, these different methods lacks of a common theoretical ground on which they can be built and compared. In fact, the analytical development of convolutional and pooling methods is not inherited from an underlying framework. Instead, these methods and their mathematical descriptions are obtained ad hoc, and they are based upon suitable heuristic observations.

In this paper, we establish a general framework which has the potential to address this crucial issue. We firstly proceed to disentangle the affine transformation $\mathcal{A}: \mathbb{R}^B \to \mathbb{R}$ from the receptive field transformation $\mathcal{R}: \mathbb{R}^N \to \mathbb{R}^B$ applied by a neuron to the input $x \in \mathbb{R}^N$, so that we can write it as
\begin{equation} \label{eq:LRsplit}
y = \mathcal{A} \left( \mathcal{R}(x) \right)
\end{equation}
Then, we formulate the receptive field by means of a set of probability density functions $\phi_i(t)$, $i=1, ..., B$, which will determine the regions of the input to be transformed by the neuron. 
However simple, we show how this approach generalizes the transformations underlying fully connected layers, convolutions, max pooling, average pooling, and min pooling. These are all particular cases of this general framework, in which they are analytically developed as a consequence of a specific choice of receptive field densities.

We will consider the case in which the input is a signal, i.e. a time series or an image. In these cases, the cardinality of the input space depends on the resolution inherent to the process which produced the data. Nonetheless, the intrinsic dimension of a data representation is, in general, independent from the resolution, and usually much lower \cite{Ansuini2019IntrinsicDO}. In our framework, the input dimension and the number of parameters of a neuron are naturally untied.

The probability density functions which defines the receptive field can be flexibly parameterized, and they can depend on the input. We show how to analytically derive the transformation $\mathcal{R}(x)$ in (\ref{eq:LRsplit}), and we demonstrate that, under mild assumptions on the set of densities, it is differentiable. We call the layers developed according to this perspective \textit{density-embedding layers}, because of the direct link between the selected densities and the transformation that they determine on the the input.

\section{The generalized neuron}

The artificial neuron applies an affine transformation to the input $ x \in \mathbb{R}^N$, which is followed by a non-linear activation. Hereafter, we consider only the affine step, and we leave implicit that it will be further transformed by a proper activation function.

In the conventional neuron, the affine step is expressed as 
\begin{equation} \label{eq:convNeur}
y = \sum_{j=1}^N w_j x_j + b
\end{equation}
where $w_j \in \mathbb{R}$, $j=1, ..., N$ and $b \in \mathbb{R}$ are its parameters. At the foundation of our work there is the idea of generalizing the affine transformation as a scalar product of functions. By appropriately defining the parameter function $w: \mathbb{R} \to \mathbb{R}$ and the input function $x: \mathbb{R} \to \mathbb{R}$ we can rewrite the previous affine transformation as:
\begin{equation} \label{eq:funcSP}
y = \int_0^N w(t) x(t) \mathrm{d}t + b
\end{equation}
In order to simplify computations, we express these functions as linear combinations $w(t) = \sum_{i=1}^B w_i \phi_i(t)$ and $x(t) = \sum_{n=1}^N x_n s_n(t)$, where $\phi_i(t)$ and  $s_n(t)$ are respectively $B$ and $N$ basis functions, and have the only property to be integrable on the interval $[0, N]$. With slight abuse of notation, we indicate the vectors of their coefficients with $w \in \mathbb{R}^B$ and $x \in \mathbb{R}^N$.

In most cases, the set of functions $s_n(t)$ will be the piecewise constant set of functions $\Delta(I_n): \mathbb{R} \to \mathbb{R}$ such that $\Delta(t; I_n) = 1$ for $t \in I_n = [n-1, n]$, and $0$ otherwise. This is because the input signal is almost always provided as a vector.
Moreover, in this more general representation the number of parameters $B$ and the cardinality of the input space $N$ are disentangled, i.e. the number of neuron's parameters can be different from the input signal resolution. Note that if we also express $w(t)$ in terms of the functions $s_n(t)$, than (\ref{eq:funcSP}) reduces to (\ref{eq:convNeur}), hence we obtain the conventional neuron as a special case.

By substituting the function expressions in (\ref{eq:funcSP}) we get:
\begin{equation} \label{eq:tauDef}
y = \sum_{i=1}^B \sum_{n=1}^N w_i x_n \int_0^N \phi_i(t) s_n(t) \mathrm{d}t + b = \sum_{i=1}^B w_i \sum_{n=1}^N \Gamma_{in} x_n + b = w^T \Gamma x + b
\end{equation}
where the matrix $\Gamma = \big(\Gamma_{in}\big)$, $\Gamma_{in} = \int_0^N \phi_i(t) s_n(t) \mathrm{d}t$ describes the interaction between the two bases of functions. We see that (\ref{eq:tauDef}) expresses an affine transformation which is similar to (\ref{eq:convNeur}). But what is the effect of $\Gamma$ on $x$? By fixing $\phi_i$ to a probability density function, the $i^{th}$ component of the resulting vector is:
\begin{equation} \label{eq:gammaEff}
\big(\Gamma x \big)_i = \sum_{n=1}^N x_n \int_0^N \phi_i(t) s_n(t) \mathrm{d}t = \int_0^N \phi_i(t) x(t) \mathrm{d}t = \mathbb{E}_{\phi_i}\big[x(t) \big]
\end{equation}
The choice of $\phi_i(t)$ as density functions, which will be enforced from now on, allows us to interpret $\Gamma x$ as the vector of the expected values of $x(t)$ with respect to the elements of the basis. Since $\Gamma$ is weighting the regions of the input that are transformed by the neuron, it is clear that its effect is to determine the receptive field of the neuron with respect to the input. In fact, $\phi_i(t)$ describes directly the shape of the receptive field. Differently from (\ref{eq:convNeur}), in (\ref{eq:tauDef}) we can find the form expressed in (\ref{eq:LRsplit}), where the receptive field action $\mathcal{R}(x) = \Gamma x$ and the affine transformation $\mathcal{A}$ are disentangled. Therefore, we can analytically prescribe the former regardless of the latter.

We notice that the densities $\phi_i(t)$ can be dependent on the input, and they can even be parameterized. Hence we can consider densities in the form $\phi_i(t, x, \lambda)$, where $\lambda$ is the vector of the density's parameters, and $x$ is the vector of coefficients of the input signal.

\subsection{An analytical expression for $\Gamma$}
With no further assumptions on the mathematical properties of the density functions $\phi_i(t, x, \lambda)$, in general we can evaluate $\Gamma$ by numerical integration. In fact, if the densities are fixed, i.e. their parameters do not change and they do not depend on the input, it is sufficient to compute $\Gamma$ only once at the initialization of the neuron. For instance, the fully connected and the convolutional layers belong to this setting. In this case the receptive field of their individual neurons is constant. Nonetheless, we are interested in a more general setting in which the densities are able to adapt with respect to the input and can be described by learnable parameters. In this case, two problems occur:
\begin{enumerate}
    \item $\Gamma$ must be numerically evaluated at every new iteration, which is computationally expensive;
    \item Since numerical integration is involved, we cannot benefit from the efficiency of the existing automatic differentiation systems, which are provided in frameworks like PyTorch and Tensorflow. 
\end{enumerate}
We now demonstrate that these issues can be addressed by conveniently choosing densities. Let $\phi_i(t, x, \lambda)$ be a Riemann integrable function on the interval $[0, N]$ for every choice of $\lambda$ and for every $x \in \mathbb{R}^N$, and let it admit a primitive $F_i(t, x, \lambda)$ expressible by means of elementary functions on the same interval. For the second fundamental theorem of calculus,
\begin{equation} \label{eq:fundThCalc}
\int_0^N \phi_i(t, x, \lambda) = F_i(N, x, \lambda) - F_i(0, x, \lambda)
\end{equation}
Moreover, we can furtherly simplify this expression by assuming the most common case in which the input is given as a vector. Therefore, we can equivalently express the input function as a piecewise constant function:
\begin{equation} \label{eq:gammaSimpl}
x(t) = \sum_{n=1}^N x_n \Delta(t; I_n), \quad \textrm{with} \quad \Delta(t; I_n) = \left\{
        \begin{array}{ll}
            1 & \quad t \in I_n = [n-1, n] \\
            0 & \quad \textrm{otherwise}
        \end{array}
    \right.
\end{equation}
From now on, we will always assume this expression for the input function. In this way, the expression for $\Gamma$ simplifies to: 
\begin{equation} \label{eq:gammaSimpl2}
\Gamma_{in} = \int_0^N \phi_i(t, x, \lambda) s_n(t) \mathrm{d}t = \int_{I_n} \phi_i(t, x, \lambda) \mathrm{d}t = F_i(n, x, \lambda) - F_i(n-1, x, \lambda)
\end{equation}
Given an analytical expression of $F$, the computation of $\Gamma$ can be performed exactly and, since $F$ is an elementary function, we can differentiate by means of any automatic differentiation system.

\subsection{Extension for images}\label{sec:extensionImages}
So far, we have considered 1D input signals, but the extension to the N-dimensional scenario is straightforward. In particular, we are interested in the case of 2D inputs, like images. In this case, $\Gamma$ becomes a 4-order tensor $\Gamma = \big(\Gamma_{ijmn}\big)$:
\begin{equation} \label{eq:gamma2D}
\Gamma_{ijmn}=\int_{I_m}\int_{I_n} \phi_{ij}(t, u, x, \lambda)\mathrm{d}t\mathrm{d}u
\end{equation}
and the receptive field action is expressed by $\mathcal{R}(x) = \sum_{m,n} \Gamma_{ijmn}x_{mn}$.

We observe that, by assuming separable densities, i. e. $\phi_{ij}(t, u, x, \lambda) = f_{ij}(t, x, \lambda) g_{ij}(u, x, \lambda)$, we obtain a further simplification:
\begin{equation} \label{eq:gamma2Dsimpl}
\Gamma_{ijmn}=\int_{I_m}\int_{I_n} f_{ij}(t, x, \lambda) g_{ij}(u, x, \lambda)\mathrm{d}t\mathrm{d}u = \int_{I_n} f_{ij}(t, x, \lambda)\mathrm{d}t   \int_{I_m} g_{ij}(u, x, \lambda)\mathrm{d}u = \Gamma_{ijn} \Gamma_{ijm}
\end{equation}
For the sake of clarity and in the light of the last expression, from now on we will consider the case of 1D input signals.

\section{Density-embedding layers}
The framework allow us to define a layer by specifying a set of density functions. We refer to the layers defined in this way as density-embedding layers. We will now show how different layers can be obtained by choosing appropriate sets of densities. Specifically, we demonstrate how the fully connected layer and the convolutional layer are recovered under this framework. Typically, the receptive field of the neurons of a layer is defined by means of hyperparameters, such as kernel size and stride. The type of pooling is usually selected by hand too. Moreover, in most transformations the receptive field covers the input uniformly. 
By suitably parameterizing the set of densities, it is possible to develop layers with a more flexible receptive field which adapts to data. We demonstrate with two simple examples how adaptive kernels and adaptive pooling can be obtained within this framework.

\subsection{The fully connected layer}
It order to build the fully connected layer, we observe that every neuron $l$ must have the same receptive field, i.e. $\mathcal{R}_l(x) = \mathcal{R}(x)$ $\forall l \in \{1, ..., L\}$. Under this framework, this translates in densities which are independent from the specific neuron, i. e. $\phi_i^l = \phi_i$. Moreover, every density function of the receptive field collects exactly one element of the input signal. Therefore, we prescribe a set of piecewise constant densities $\Delta(I_i)$ defined on the partition of intervals $I_i = I_n$, with $i, n =1, ..., N$, where $I_n = [n-1, n]$ represents the natural partition of the input. The $\Gamma_{in}$ element is computed as
\begin{equation} \label{eq:fullyConnected}
\Gamma_{in} = \int_{I_n} \phi_i(t, x, \lambda) \mathrm{d}t = \int_{I_n} \Delta(t; I_n) \mathrm{d}t = \left\{
        \begin{array}{ll}
            1 & \quad \textrm{if} \quad i = n \\
            0 & \quad \textrm{otherwise}
        \end{array}
    \right.
\end{equation}
As expected, $\Gamma$ is the identity, so that we get $y = w^T\Gamma x + b = w^Tx + b$.

\subsection{The convolutional layer}
Let us consider a 1D convolutional layer, where the stride is set to $S$ and the kernel size to $K$. The receptive field of the $l^{th}$ neuron of this layer covers uniformly the $K$ elements of the input corresponding to set of intervals $I_i^l = [(l-1) S + i -1, (l-1)S + i]$, $i=1, ..., K$. Every interval covers a precise element of the input. Therefore, the receptive field densities of the $l^{th}$ neuron are $\Delta(t; I_i^l)$, and we get
\begin{equation} \label{eq:conv}
\Gamma_{in}^l = \int_{I_n} \Delta(t; I_i^l)(t)\mathrm{d}t= \left\{
        \begin{array}{ll}
            1 & \quad \textrm{if} \quad n = (l-1)S + i \\
            0 & \quad \textrm{otherwise}
        \end{array}
    \right.
\end{equation}
Every neuron has a sparse receptive field, i. e. its densities only cover a small region of the input signal, and this region is constant. By sharing the same set of $K$ weights (i.e. the kernel) among the neurons of a layer, we recover one channel of a convolutional layer. We can obtain different channels by associating different kernels to the same set of $L$ matrices $\Gamma^l$.

\subsection{Adaptive convolution}
We extend the the receptive field densities in the last example to a parameterized form. For the sake of simplicity, we will still use a set of uniform distributions, but we define them on the intervals
\begin{equation} \label{eq:adaConvIntervals}
I_i^l(p) = \left[ (l-1)S + \frac{p}{K}(i -1), (l-1)S + \frac{p}{K}i \right]
\end{equation}
where $p \in \mathbb{R}$ is the \textit{kernel amplitude}, and it will be learned by gradient descent. It defines the extension of the local receptive field of a neuron over the input, and in this example it is shared among all the neurons of the layer.
By considering that, for any set of reals $a_1, b_1, a_2, b_2$,
\begin{equation} \label{eq:formula1}
\int_{[a_1, b_1]} \Delta(t; [a_2, b_2])\mathrm{d}t = \int_{[a_1, b_1] \cap [a_2, b_2]} \mathrm{d}t = \mathrm{max}(0, \mathrm{min}(b_1, b_2) - \mathrm{max}(a_1,a_2))
\end{equation}
we can easily compute the elements of $\Gamma$ for the adaptive convolution:
\begin{equation} \label{eq:gammaAdaConv}
\begin{split}
\Gamma_{in}^l(p) = \int_{I_n} \Delta(t; I_i^l(p))\mathrm{d}t = \quad \quad \quad \quad \quad \quad \quad \quad \quad \quad \quad \quad  \\
 = \mathrm{max}\left(0, \mathrm{min}\left(n, (l-1)S + \frac{p}{K}i\right) - \mathrm{max}\left(n-1, (l-1)S + \frac{p}{K}(i -1)\right)\right)
\end{split}
\end{equation}
where, as before, $I_n = [n-1, n]$.
$\Gamma$ can be automatically differentiated with respect to $p$, hence $p$ can be learned. In the traditional convolution, the kernel amplitude $p$ is equal to the kernel size $K$, i.e. the extension of the region of the input covered by the kernel is equal to the number of parameters of the kernel. Specifying this simple parameterization makes the kernel able to expand or contract and, eventually, to adapt its amplitude to the specific features of the data.

More important is that this is just one of many possible parameterization which could be inspected under this framework, and which could differ for effectiveness and robustness. For instance, the stride could be parameterized, or it could be defined to be proportional to the kernel amplitude $p$. In addition, a function $p(\lambda)$ could be used instead of $p$. For example, one method to bound the range of values of the kernel amplitude to an interval $[a, b]$ is to define $p(\lambda) = a + (b-a)\sigma(\lambda)$, where $\sigma$ is the logistic function and $\lambda$ is the real parameter to be learned instead of $p$.

\subsection{Adaptive pooling}
Defining a pooling operation means defining a receptive field, and many techniques have been proposed to adapt the pooling operation to data \cite{Kobayashi2019GaussianBasedPF, Lee2018GeneralizingPF, Saeedan2018DetailPreservingPI, Kobayashi2019GlobalFG}. One way to achieve it consists in parameterizing the pooling operation by means of a learnable real parameter. Starting from a specific set of densities, we show how an adaptive pooling technique can be obtained. We parameterize the set of densities by means of a parameter $\beta \in \mathbb{R}$, and we obtain a receptive field which is able to reproduce the max pooling ($\beta \to \infty$), the average pooling ($\beta = 0$), and the min pooling ($\beta \to -\infty$). Similarly to the convolutional example, we select a set of intervals defining specific regions of the input signal, and we use a set of densities which are uniform over the intervals to compute the $\Gamma$ matrix. However, we assign a general parameterization to the intervals to indicate that their features (length, position, etc.) can be learned by gradient descent. One additional core difference with respect to the previous examples is that in this setting the densities depends on the input $x \in \mathbb{R}^N$.

Let us consider a set of intervals over the input's domain $J_i(\lambda) \subset [0, N]$, $i=1, ..., B$, where $\lambda$ is a generic set of parameters describing arbitrary interval features. For instance, the intervals $I(p)$ introduced in the adaptive pooling are an example of parameterized intervals. Given the usual input signal $x(t) = \sum_{n=1}^N x_n \Delta(t; I_n)$, we define the receptive field densities as
\begin{equation} \label{eq: RFAdaPool}
\phi_i(t, x, \beta, \lambda) = \Delta(t; J_i(\lambda)) \,\exp\left( \beta \, \Delta(t; J_i(\lambda)) \, x(t) \right) \left( \int_{J_i} \exp\left( \beta \, \Delta(t; J_i(\lambda)) \, x(t) \right) \mathrm{d}t \right)^{-1}
\end{equation}
The result of the integration of the density $\phi_i(t, x, \beta, \lambda)$ on the interval $I_n = [n-1, n]$ is
\begin{equation} \label{eq: gammaAdaPool}
\Gamma_{in}(x, \beta, \lambda) = \frac{m_{in} e^{\beta x_n}}{\sum_{r=1}^N m_{ir} e^{\beta x_r}}
\end{equation}
where $m_{in} = \int_{I_n \cap J_i} \mathrm{d}t$ (see supplementary material for further details on the mathematical steps). 

Notice that $\left( \Gamma(x, \beta, \lambda) \, x \right)_i = \sum_{n=1}^N \Gamma_{in} \, x_n$ is the $i^{th}$ output of a max pooling, average pooling, and min pooling transformation respectively for $\beta \to \infty$, $\beta=0$, and $\beta \to -\infty$.

\section{Experimental results}
Density-embedding layers constitute a very broad family of transformations, and their formulation allow to flexibly shape the receptive field that will select the input regions to be forwarded to the next layer. Hereafter, we show two implementations of density-embedding layers based on the logistic distribution. To highlight their properties and to provide a visualization, we build two very simple networks which are constituted only by those layers. In order to show that they can provide a representation of the input which is more parsimonious but still accurate, we compare their performances with two fully-connected networks on MNIST dataset. MNIST dataset has a training set of 60,000 examples, and a test set of 10,000 examples, where every input is a $28 \times 28$ image with a single channel.

We implemented the density-embedding layers with PyTorch and tested them on the MNIST dataset. We compared their performances with simple fully-connected neural networks. Every model has been trained for 20 epochs through Adam optimization, with a maximum learning rate of 0.002, and this process has been repeated for 5 runs with random initializations. All the models were trained on NVIDIA GeForce GTX 1050, and the results are shown in Table \ref{table:1}. All the additional operations involving the computation of the tensor $\Gamma$ can be efficiently parallelized and automatically differentiated. Further details can be found in the supplementary material.

\subsection{Logistic-embedding layer}
The logistic distribution is defined as
\begin{equation} \label{eq: gammaAdaPool2}
f(t; \mu, s) = \frac{e^{-(t-\mu)/s}}{s\left(1 + e^{-(t-\mu)/s}\right)^2}
\end{equation}
where $\mu$ is the mean and the variance is given by $\frac{s^2 \pi^2}{3}$.
This distribution approximates well a gaussian distribution, but it has the considerable advantage of having a cumulative distribution which can be expressed by means of elementary functions. Specifically, its cumulative distribution is the logistic function, which is written as
\begin{equation} \label{eq: gammaAdaPool3}
F(t; \mu, s) = \frac{1}{1 + e^{-(t-\mu)/s}}
\end{equation}
We use this density function to build a layer for processing images.

For the sake of simplicity, let us consider single-channel input images $x \in \mathbb{R}^{N \times N}$, where we indicate the element $(m, n)$ of the image with $x_{mn}$, $m,n=1,...,N$. We define the set of densities $\phi_{ij}$, $i,j=1,...,B$ as
\begin{equation} \label{eq: logisticrf}
\phi_{ij}(t, u; \alpha_{ij}, \beta_{ij}) = f(t; \mu(\alpha^{(1)}_{ij}), s(\beta^{(1)}_{ij}))\,f(u; \mu(\alpha^{(2)}_{ij}), s(\beta^{(2)}_{ij}))
\end{equation}
where $\alpha_{ij} = (\alpha^{(1)}_{ij}, \alpha^{(2)}_{ij}) \in \mathbb{R}^2,\, \beta_{ij} = (\beta^{(1)}_{ij}, \beta^{(2)}_{ij}) \in \mathbb{R}^2$ are parameters to be learned by gradient descent. Therefore, $\phi_{ij}$ is a set of $B \times B$ two-dimensional density functions obtained by the product of two logistic distributions with different $\mu$ and $s$. For every density, $\mu$ and $s$ are the following functions of learnable parameters:
\begin{equation} \label{eq: mus}
\mu(z) = \frac{N}{1 + e^{-z}},\quad s(z) = \frac{N}{1 + e^{-z}}.
\end{equation}
Although we could directly learn $\mu$ and $s$ for every density, we actually learn $\alpha_{ij}$ and $\beta_{ij}$ to restrict $\mu$ and $s$ on the $[0, N]$ interval, i.e. on the input's domain. Note that we used a logistic function to express parameters, but any bounded function can be employed. Therefore, every density function $\phi_{ij}$ is described by four parameters, and we have a total of $4 \times B \times B$ parameters. 
According to the methodology described in Section \ref{sec:extensionImages}, we obtain $\mathcal{R}(x; \alpha, \beta) = \Gamma(\alpha, \beta)\, x$, where $\mathcal{R}: \mathbb{R}^{N \times N} \to \mathbb{R}^{B \times B}$ is a linear function of the input, and we call it \textit{logistic-embedding layer}.

Every logistic-embedding layer learns a set of $B \times B$ logistic distributions, which are used to apply a pooling on the input image. The result is a $B \times B$ filter, where every pixel represent the expected value with respect to one of the distributions. The filter is then flattened and fed as input to a linear classifier. We used four different values of $B$ (3, 5, 8, 15) and compared them with a fully connected linear layer (linear classifier). For $B=15$, the logistic-EL reaches $8.48\%$ test error, against $7.34\%$ of the fully connected, but it saves more than half of the parameters. Moreover, for $B=8$ it exceeds $90\%$ of test accuracy with almost one tenth of the parameters of the fully connected layer.

\subsection{Learning the density parameters by microNN}
In this section, we consider the same set of logistic densities of the last example, but rather than expressing their mean as a logistic function of parameter $\alpha \in \mathbb{R}^{B \times B \times 2}$,  we use a linear micro network (mNN) to force a dependency on the input.
%
%
Therefore, we obtain a density-embedding layer where the receptive field adapts to the given input by means of a smaller network $\alpha(x;W)$, i.e. 
\begin{equation} \label{eq: gammaAdaPool4}
\mathcal{R}(x; W, \beta) = \Gamma\left(\alpha(x;W), \beta\right)\, x
\end{equation}
where $W$ indicates the parameters of the micro network.
As already shown, the performance of the logistic-embedding layer is comparable to the fully connected one, but it is significantly more parsimonious with respect to the number of parameters. For this reason, we utilized the layer described in the previous paragraph to represent the micro network $\alpha(x;W)$. The Logistic-EL used as micro network makes use of $B_0 \times B_0$ density functions for computing $\alpha$. The outer Logistic-EL will use the output of the micro network as parameters of its $B \times B$ density functions, determining the final output.

\begin{figure}
\centering
\begin{subfigure}[b]{.45\textwidth}
    \includegraphics[width=\textwidth]{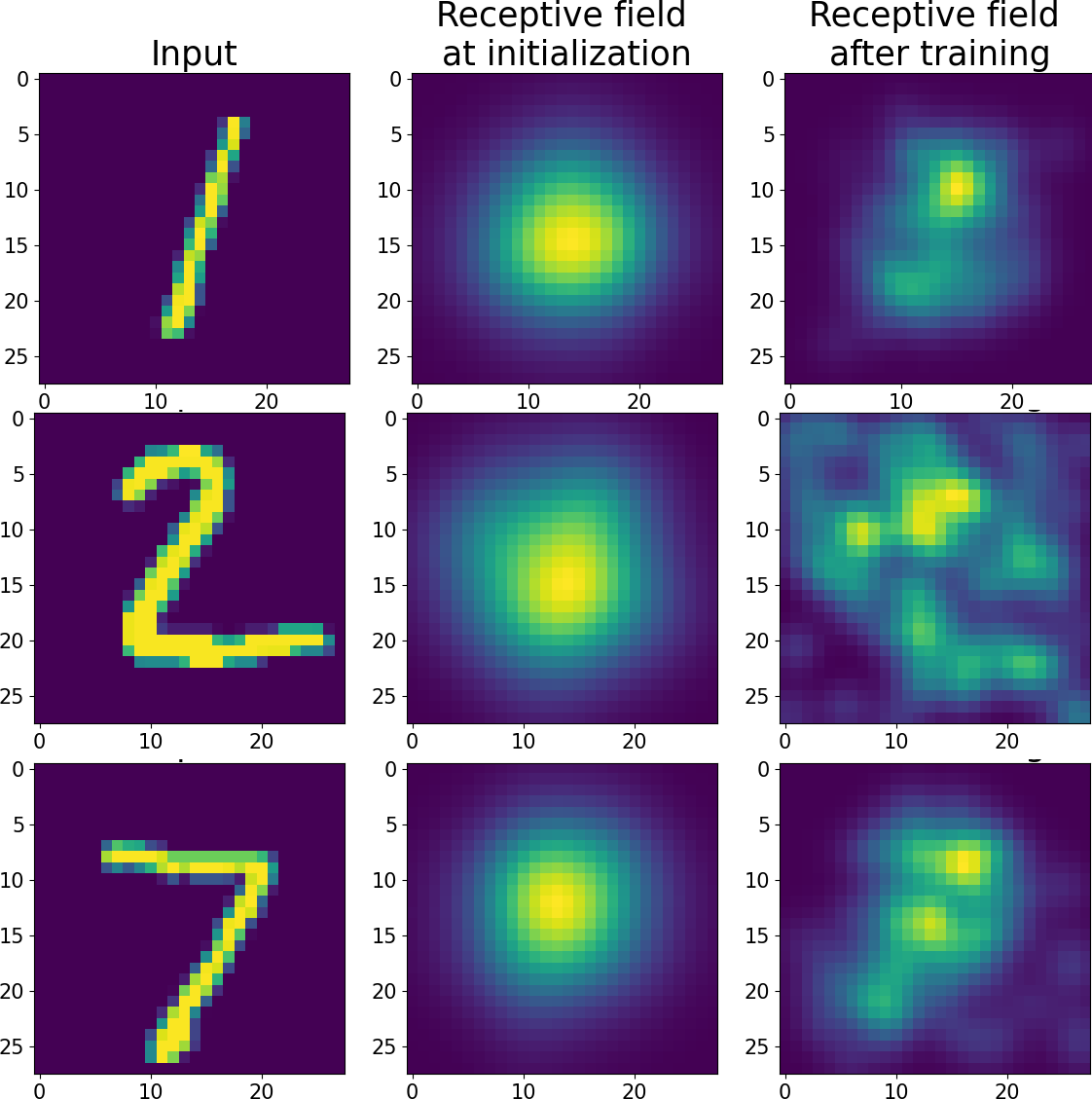} 
    \caption{}
    \label{fig:subim1}
\end{subfigure}\qquad
\begin{subfigure}[b]{.4\textwidth}
    \includegraphics[width=\textwidth]{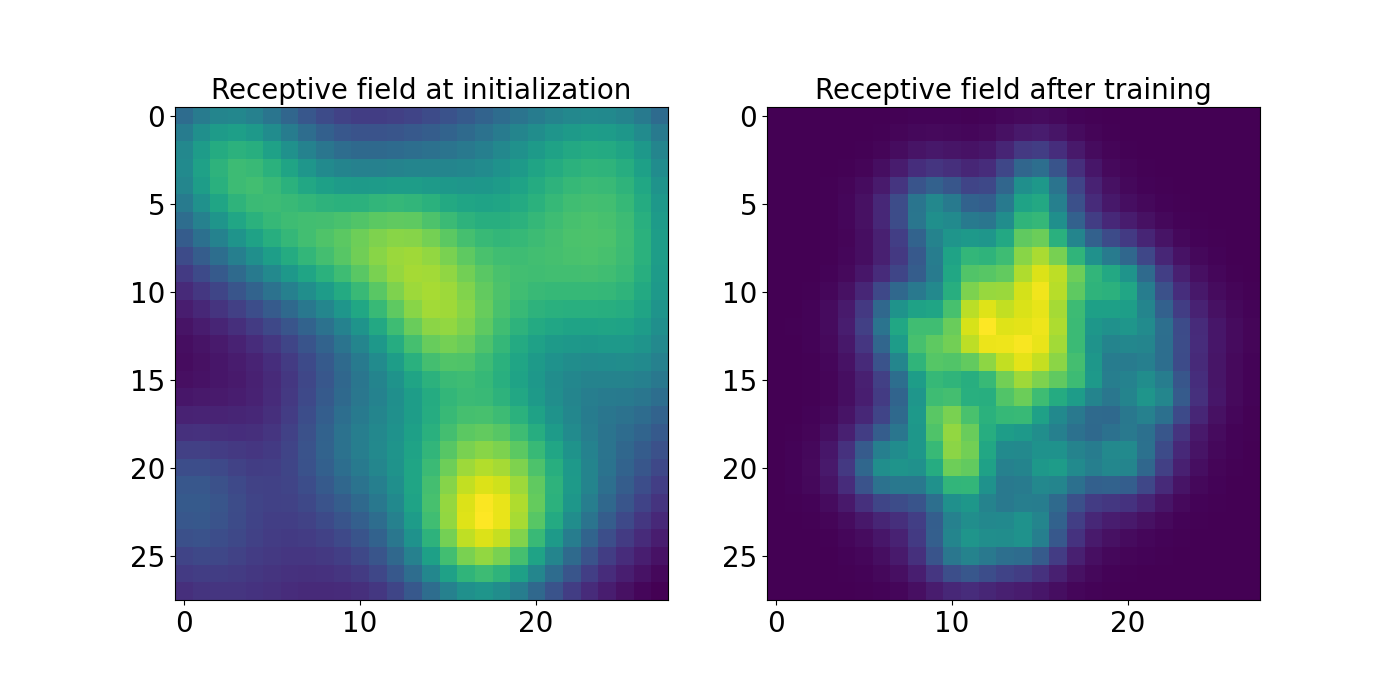}
    \caption{}
    \vspace{1ex}
    \includegraphics[width=\textwidth]{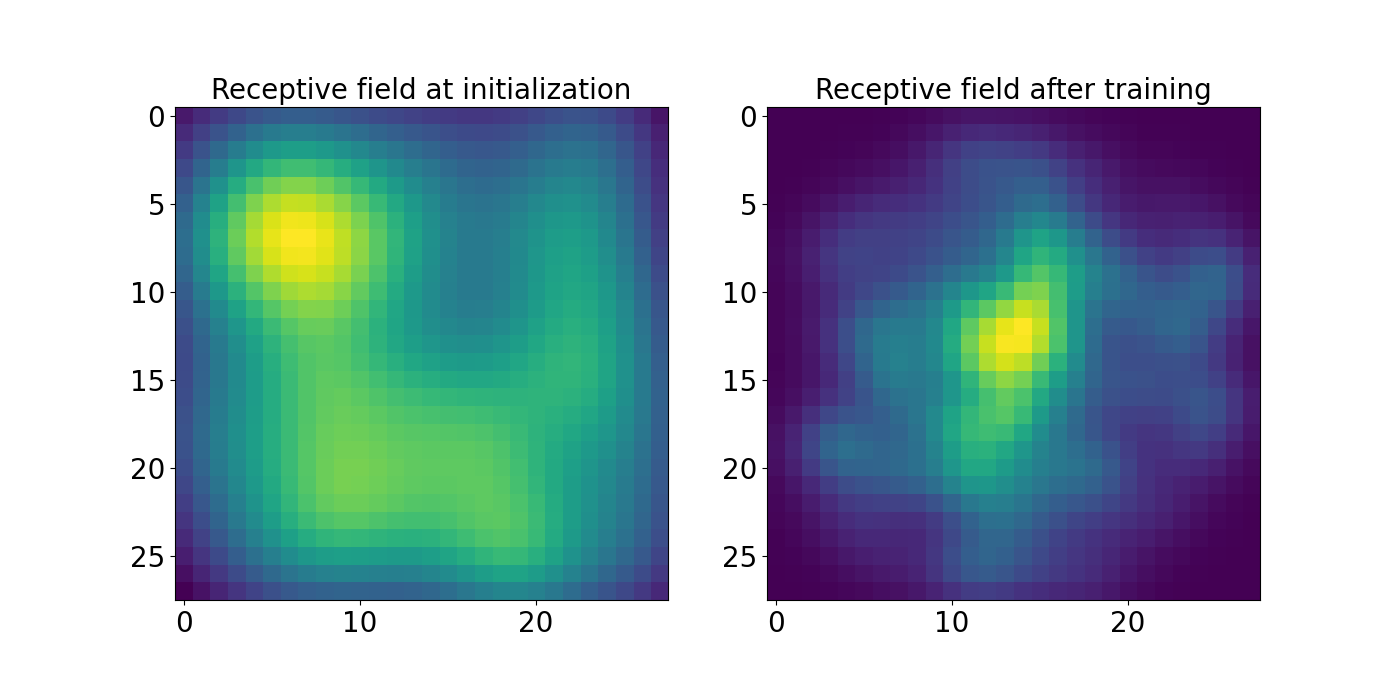}
    \caption{}
\end{subfigure}

\caption{Visualization of the receptive field before and after training. Every receptive field represents the sum of the density functions over the input domain. (a) Receptive fields of a logistic-EL with micro network ($B=10$, $B_0=5$) before (center) and after (right) training for three different inputs (left). The micro network makes the receptive field to adapt to the input. (b) and (c) show the receptive fields of a logistic-EL before (left) and after (right) training respectively for $B=5$ and $B=15$. They are fixed, since their parameters do not depend on the input.}
\label{fig:image1}
\end{figure}

Notice that, even if we compute $\alpha$ by means of a linear function, the full layer is not linear with respect to the input. In fact, the output of the Logistic-EL is a nonlinear function of its parameters $\alpha$ and $\beta$. Since $\alpha$ depends on the input, we are actually applying a nonlinear transformation to $x$. For this reason, we compared this model with a fully connected network (FCN) with one hidden layer of $50$ neurons. We used three different values for $B$ (6, 8, 10), and $B_0$ was chosen equal to $B/2$.  The Logistic-LE with micro network, for $B=10$ and $B_0=5$, performs slightly better than the FCN, with a sensible reduction in the number of parameters ($6510$ against $39760$). The results are displayed in Table~\ref{table:1}, while Figure~\ref{fig:image1} visually depicts the receptive fields for the logistic-embedding layer.

\begin{table}[ht]
\centering
 \begin{tabular}{c c c c} 
 \hline
 Model & Error ($\%$)($5$ runs) & \# parameters \\ [0.5ex] 
 \hline\hline
 FC (no hidden layer) & $7.34 \pm\, 0.09$ &    $7850$ \\ 
 FC ($1$ hidden layer, $50$ neurons) &  $2.96 \pm\, 0.23$  &    $39760$ \\ 
 Logistic-EL ($B=3$) &    $17.80 \pm\, 0.93$ &    $136$ \\
 Logistic-EL ($B=5$)  &   $10.52 \pm\, 0.11$ &    $360$ \\
 Logistic-EL ($B=8$)  &   $9.13 \pm\, 0.15$ &     $906$ \\
 Logistic-EL ($B=15$)  &  $8.48 \pm\, 0.17$  &    $3160$ \\ 
 Logistic-EL with mNN ($B=6$, $B_0=3$) & $4.44 \pm\, 0.13$ &     $1198$ \\
 Logistic-EL with mNN ($B=8$, $B_0=4$) & $3.39 \pm\, 0.23$ &     $3018$ \\
 Logistic-EL with mNN ($B=10$, $B_0=5$) & $\textbf{2.91} \pm\, \textbf{0.16}$ &     $6510$ \\ [1ex]
 \hline
 \end{tabular}
 \caption{Performance comparison on MNIST}
\label{table:1}
\end{table}

\section{Conclusions}

In this work we proposed a novel general framework for defining a broad category of layers of neurons by explicitly representing the receptive field with a set of density functions. We have shown that these density functions can be selected and parameterized  flexibly, under the only  condition that their primitive can be expressed by means of elementary functions. Moreover, they are able to depend on the input in nontrivial ways. We have shown how  our approach  recovers the fully connected and the convolutional layers as particular cases, and we have developed further examples to show how adaptive differentiable layers can be naturally described.

Finally, we have developed two variants of a density-embedding layer based on the logistic distributions, and we have demonstrated how they are able to learn receptive fields which effectively leverage on the input properties and allow to significantly reduce the number of parameters.

It is important to mention that the logistic-embedding layer is one of many possible density-embedding layers which deserve to be explored. The value of this framework lies in the way it allows to directly shape the receptive field of artificial neurons. The receptive field determines what information about the input is forwarded and elaborated, and selecting it cleverly is crucial for generalization and memory efficiency. We believe that this methodology is a convenient tool for studying new adaptive layers, and it represents a potential candidate as a theoretical framework for analytically comparing the properties of a rich family of transformations. Future work involves the exploration of different sets of densities, and a broader experimental analysis and validation of the properties of different receptive fields.

\section*{Broader Impact}
As our proposals consists in a theoretical framework, we believe that the impact of our work on social and ethical aspects can only be indirect. Within this framework, we can develop layers which significantly reduce the number of parameters required in the fully connected layer. Through further investigations, we hope to be able to derive efficient and scalable models to be used in a broad spectrum of problems. These applications can have an impact on social and ethical issues.

\bibliographystyle{plain} 
\bibliography{main} 

\newpage

\appendix
\section{Appendix}

\subsection{Adaptive pooling}
Let us consider a set of intervals over the input's domain $J_i(\lambda) \subset [0, N]$, $i=1, ..., B$, where $\lambda$ is a generic set of parameters describing arbitrary interval features. Given the input signal $x(t) = \sum_{n=1}^N x_n \Delta(t; I_n)$, we define the receptive field densities as
\begin{equation} \label{eq: RFAdaPool2}
\phi_i(t, x, \beta, \lambda) = 
\Delta(t; J_i(\lambda)) \,\exp\left( \beta \, \Delta(t; J_i(\lambda)) \, x(t) \right) 
\left( \int_{J_i} \exp\left( \beta \, \Delta(t; J_i(\lambda)) \, x(t) \right) \mathrm{d}t \right)^{-1}
\end{equation}
Hereafter, we show the details of the integration of $\phi_i(t, x, \beta, \lambda)$ on the interval $I_n = [n-1, n]$, $n\in \{1,...,N\}$ where $N$ is a natural number indicating the dimension of the input space.

Let us first consider an arbitrary interval $A \subseteq \bigcup_{n=1}^N I_n = [0, N]$.
\begin{equation} \label{eq: RFAdaPool3}
\begin{split}
\int_A e^{\beta \, \Delta(t; A) \, x(t)} \mathrm{d}t = 
\int_A e^{\beta \, \Delta(t; A) \, \sum_{n=1}^N x_n \Delta(t; I_n)} \mathrm{d}t
\end{split}
\end{equation}
Since the integration domain is $A$ and $\Delta(t; A) = 1\, \forall t \in A$, we get:
\begin{equation} \label{eq: third}
\begin{split}
\int_A e^{\beta \, \Delta(t; A) \, \sum_{n=1}^N x_n \Delta(t; I_n)} \mathrm{d}t = 
\int_A e^{\beta \, \sum_{n=1}^N  x_n \,\Delta(t; I_n)} \mathrm{d}t
\end{split}
\end{equation}
Since $A$ is a subset of $[0, N]$, the last expression can be rewritten as
\begin{equation} \label{eq: third2}
\begin{split}
\int_0^N \Delta(t; A)\, e^{\beta \, \sum_{n=1}^N  x_n \,\Delta(t; I_n)} \mathrm{d}t = 
\sum_{n=1}^N \int_{I_n} \Delta(t; A)\,e^{\beta \, \sum_{n=1}^N  x_n \,\Delta(t; I_n)} \mathrm{d}t =\\
=\sum_{n=1}^N \int_{I_n} \Delta(t; A)\,e^{\beta \, x_n} \mathrm{d}t =
\sum_{n=1}^N e^{\beta \, x_n} \int_{I_n} \Delta(t; A) \mathrm{d}t = 
\sum_{n=1}^N e^{\beta \, x_n} \int_{I_n \cap A} \mathrm{d}t
\end{split}
\end{equation}
Therefore, 
\begin{equation} \label{eq: RFAdaPool5}
\begin{split}
\int_A e^{\beta \, \Delta(t; A) \, x(t)} \mathrm{d}t = 
\sum_{n=1}^N e^{\beta \, x_n} \int_{I_n \cap A} \mathrm{d}t
\end{split}
\end{equation}
By using this relation, we can easily compute $\Gamma_{in}$ for the adaptive pooling scenario described in section 3.4:
\begin{equation} \label{eq: RFAdaPool6}
\begin{split}
\Gamma_{in}(x, \beta, \lambda) = \int_{I_n} \phi_i(t, x, \beta, \lambda)\,\mathrm{d}t = 
\frac{
\int_{I_n}\Delta(t; J_i(\lambda)) \,e^{\beta \, \Delta(t; J_i(\lambda)) \, x(t)}\,\mathrm{d}t}
{\int_{J_i} e^{ \beta \, \Delta(t; J_i(\lambda)) \, x(t)}\mathrm{d}t
} =\\
=\frac{
\int_{I_n\cap J_i(\lambda)} \,e^{\beta \, \Delta(t; J_i(\lambda)) \,  x(t)} \,\mathrm{d}t}
{\int_{J_i(\lambda)} e^{ \beta \, \Delta(t; J_i(\lambda)) \, x(t)}\mathrm{d}t
} =
\frac{
\int_{I_n\cap J_i(\lambda)} \,e^{\beta \, \Delta(t; J_i(\lambda)\cap I_n) \, x(t)} \,\mathrm{d}t}
{\sum_{r=1}^N e^{\beta \, x_r} \int_{I_r \cap J_i(\lambda)} \mathrm{d}t
} =\\
=\frac{
\int_{I_n\cap J_i(\lambda)} \,e^{\beta \, \Delta(t; J_i(\lambda)\cap I_n) \, x(t)} \,\mathrm{d}t}
{\sum_{r=1}^N e^{\beta \, x_r} \int_{I_r \cap J_i(\lambda)} \mathrm{d}t
} =
\frac{
\sum_{r=1}^N e^{\beta \, x_r} \int_{I_n \cap I_r \cap J_i(\lambda)} \mathrm{d}t}
{\sum_{r=1}^N e^{\beta \, x_r} \int_{I_r \cap J_i(\lambda)} \mathrm{d}t
} =\\
= \frac{
e^{\beta \, x_n} \int_{I_n \cap J_i(\lambda)} \mathrm{d}t}
{\sum_{r=1}^N e^{\beta \, x_r} \int_{I_r \cap J_i(\lambda)} \mathrm{d}t
} = 
\frac{m_{in} e^{\beta x_n}}{\sum_{r=1}^N m_{ir} e^{\beta x_r}} \quad \quad \quad \quad \quad \quad \quad \quad \quad 
\end{split}
\end{equation}
where $m_{in} = \int_{I_n \cap J_i} \mathrm{d}t$.

\subsection{The expression of $\Gamma$ for the logistic-embedding layer}
Let us consider input images $x \in \mathbb{R}^{N \times N}$, where we indicate the element $(m, n)$ of the image with $x_{mn}$, $m,n=1,...,N$. As described in Section 4.1, the se of densities $\phi_{ij}$, $i,j=1,...,B$ are
\begin{equation} \label{eq: logisticrf2}
\phi_{ij}(t, u; \alpha_{ij}, \beta_{ij}) = f(t; \mu(\alpha^{(1)}_{ij}), s(\beta^{(1)}_{ij}))\,f(u; \mu(\alpha^{(2)}_{ij}), s(\beta^{(2)}_{ij}))
\end{equation}
where $\alpha_{ij} = (\alpha^{(1)}_{ij}, \alpha^{(2)}_{ij}) \in \mathbb{R}^2,\, \beta_{ij} = (\beta^{(1)}_{ij}, \beta^{(2)}_{ij}) \in \mathbb{R}^2$. Even if $\mu$ and $s$ could be learned directly, to assure that they are both bounded in the interval $[0, N]$ we learn their logits $\alpha$ and $\beta$. More specifically, we computed $\mu$ and $s$ as
\begin{equation} \label{eq: mus2}
\mu(z; p_{\mu}, q_{\mu}) = q_{\mu} + \frac{N}{1 + e^{-p_{\mu}z}},\quad s(z; p_s, q_s) = q_s + \frac{N}{1 + e^{-p_s z}}.
\end{equation}
We set $(p_{\mu}, q_{\mu}) = (4, 0)$ and $(p_s, q_s) = (0, 1)$. 

The function $f$ is the expression of the logistic distribution, and its cumulative distribution is the logistic function $F$:
\begin{equation} \label{eq: gammaAdaPooltw}
f(t; \mu, s) = \frac{e^{-(t-\mu)/s}}{s\left(1 + e^{-(t-\mu)/s}\right)^2}, \quad F(t; \mu, s) = \frac{1}{1 + e^{-(t-\mu)/s}}
\end{equation}

We compute $\Gamma$ as shown in section 2.2:
\begin{equation} \label{eq: gammaAdaPooltr}
\begin{split}
\Gamma_{ijmn}=\int_{I_m}\int_{I_n} \phi_{ij}(t, u; \alpha_{ij}, \beta_{ij}) \mathrm{d}t\mathrm{d}u = \quad \quad \quad \quad \quad \quad \quad \\ = \int_{I_n} f(t; \mu(\alpha^{(1)}_{ij}), s(\beta^{(1)}_{ij}))\mathrm{d}t   \int_{I_m} f(u; \mu(\alpha^{(2)}_{ij}), s(\beta^{(2)}_{ij}))\mathrm{d}u = \Gamma_{ijn} \Gamma_{ijm}
\end{split}
\end{equation}
where $I_n = [n, n-1]$ and $I_m = [m, m-1]$. The computation of both $\Gamma_{ijn}$ and $\Gamma_{ijm}$ is trivial, since we have the primitive of $f$:
\begin{equation} \label{eq: gammaAdaPool7}
\begin{split}
\Gamma_{ijn} = F(n; \mu(\alpha^{(1)}_{ij}), s(\beta^{(1)}_{ij}))) - F(n-1;  \mu(\alpha^{(1)}_{ij}), s(\beta^{(1)}_{ij}))) \\ 
\Gamma_{ijm} = F(m; \mu(\alpha^{(2)}_{ij}), s(\beta^{(2)}_{ij}))) - F(m-1;  \mu(\alpha^{(2)}_{ij}), s(\beta^{(2)}_{ij})))
\end{split}
\end{equation}

To generalize this expression to the case of images with $C$ channels, we can either choose to learn a different receptive field for every channel or share the same receptive field for all the channels. The former case is obtained by define a set of densities for every channel $c=1, ..., C$:
\begin{equation} \label{eq: logisticrfte}
\phi_{cij}(t, u; \alpha_{cij}, \beta_{cij}) = f(t; \mu(\alpha^{(1)}_{cij}), s(\beta^{(1)}_{cij}))\,f(u; \mu(\alpha^{(2)}_{cij}), s(\beta^{(2)}_{cij}))
\end{equation}
so that we obtain a tensor $\Gamma_{cijmn}$ and we compute the receptive field on $x_{cmn}$ as $\mathcal{R}(x) = \left(\sum_{mn}\Gamma_{cijmn}\, x_{cmn}\right) \in \mathbb{R}^{C \times B \times B}$. The other option is to share the same receptive fields for all the channels, where we can compute $\Gamma_{ijmn}$ as described above, and the receptive field on $x_{cmn}$ is simply given by  $\mathcal{R}(x) = \left(\sum_{mn}\Gamma_{ijmn}\, x_{cmn}\right) \in \mathbb{R}^{C \times B \times B}$.

We highlight that performing tensor-tensor multiplication in PyTorch is extremely straightforward, thanks to the method \texttt{torch.einsum()}.

\subsection{Initialization of the logistic-embedding layer}
The parameters of every logistic-embedding layer, i.e. the logits $\alpha_{ij}$ and $\beta_{ij}$, have been initialized according to the normal distributions $\mathcal{N}(0, 0.4)$ and $\mathcal{N}(-3, 0.3)$ respectively.

\subsection{Training times comparison}
\begin{figure}[!ht]
\includegraphics[width=1\linewidth]{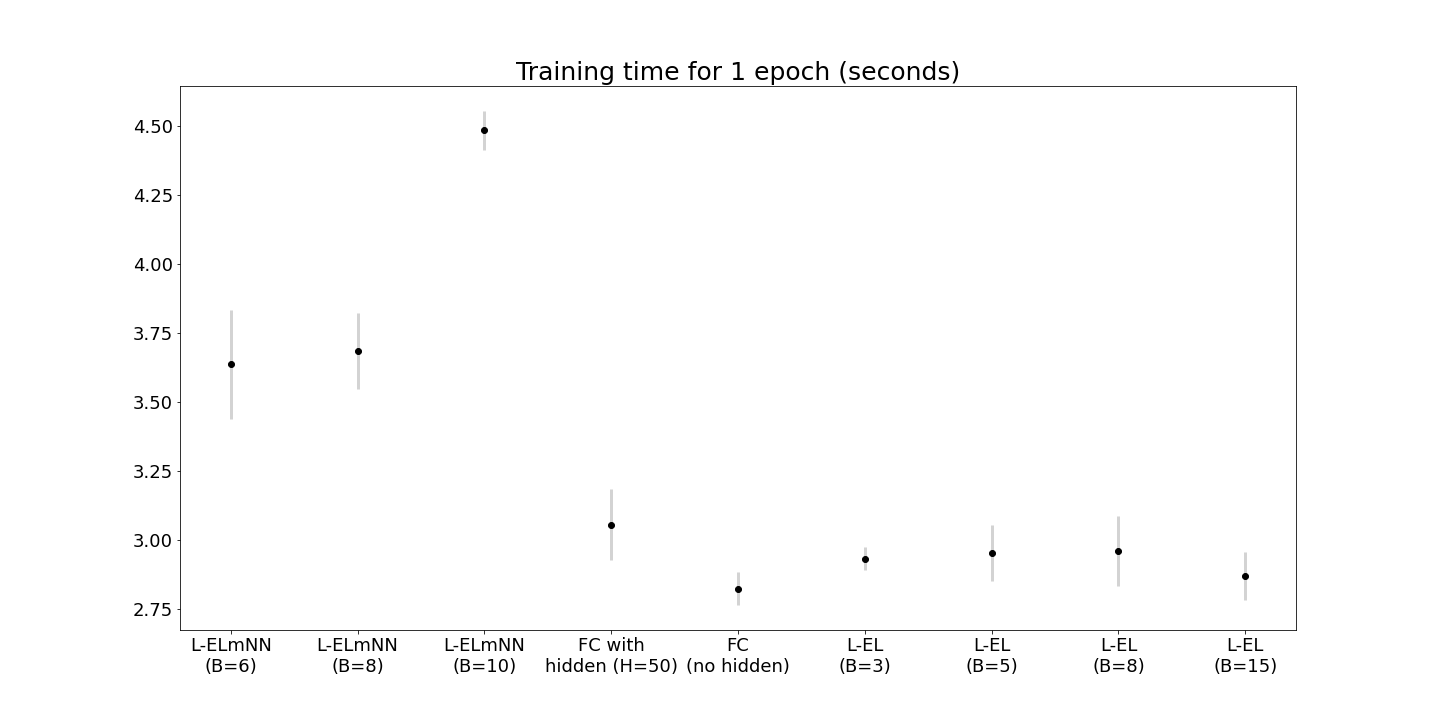}
\caption{Training times comparison. Mean and standard deviation have been computed on 10 runs.}
\label{fig:compTimes}
\end{figure}

We measured the average time for training each model for one epoch on MNIST dataset. Means and standard deviations have been computed on ten runs, under the experimental conditions described in section 4. The results are reported on Figure~\ref{fig:compTimes}.

\subsection{Experiment on CIFAR10}
We collected results on CIFAR10 dataset by means of the same models and training setting described in section 4. The results are reported in Table~\ref{table:2}.

\begin{table}[!ht]
\centering
 \begin{tabular}{c c c c} 
 \hline
 Model & Error ($\%$)($5$ runs) & \# parameters \\ [0.5ex] 
 \hline\hline
 FC (no hidden layer) & $65.06 \pm\, 0.44$ &    $30730$ \\ 
 FC ($1$ hidden layer, $50$ neurons) &  $51.80 \pm\, 0.50$  &    $154160$ \\ 
 Logistic-EL ($B=3$) &    $60.87 \pm\, 0.18$ &    $388$ \\
 Logistic-EL ($B=5$)  &   $59.55 \pm\, 0.17$ &    $1060$ \\
 Logistic-EL ($B=8$)  &   $59.34 \pm\, 0.27$ &     $2698$ \\
 Logistic-EL ($B=15$)  &  $59.61 \pm\, 0.64$  &    $9460$ \\ 
 Logistic-EL with mNN ($B=6$, $B_0=3$) & $50.97 \pm\, 0.44$ &     $7462$ \\
 Logistic-EL with mNN ($B=8$, $B_0=4$) & $\textbf{50.07} \pm\, \textbf{0.44}$ &     $21322$ \\
 Logistic-EL with mNN ($B=10$, $B_0=5$) & $50.71 \pm\, 0.22$ &     $49510$ \\ [1ex]
 \hline
 \end{tabular}
 \caption{Performance comparison on CIFAR10.}
\label{table:2}
\end{table}

\end{document}